\documentclass{article}

\usepackage{microtype}
\usepackage{graphicx}
\usepackage{subfigure}
\usepackage{booktabs} 

\usepackage{hyperref}

\newcommand{\eg}{\textit{e.g.}\xspace}
\newcommand{\ie}{\textit{i.e.}\xspace}
\newcommand{\wrt}{{w.r.t.}\xspace}

\usepackage[accepted]{icml2020}

\usepackage{amsmath,amssymb,amsfonts,latexsym,dsfont,xspace,mathtools}
\usepackage{physics,nicefrac}
\usepackage{tikz}
\usetikzlibrary{bayesnet}
\usepackage{comment}
\usepackage{float}
\usepackage{todonotes}

\newcommand{\inv}[1]{{#1}^{-1}}
\newcommand{\jac}[2]{J_{#1}(#2)}
\newcommand{\djac}[2]{\left| \det \jac{#1}{#2} \right|}

\DeclareMathOperator{\ELBO}{ELBO}
\DeclareMathOperator{\KL}{KL}
\DeclareMathOperator*{\argmax}{\arg\,\max}

\DeclareMathOperator{\Bern}{Bern}

\icmltitlerunning{Latent Transformations for Discrete-Data Normalising Flows}

\begin{document}

\twocolumn[
\icmltitle{Latent Transformations for Discrete-Data Normalising Flows}



\icmlsetsymbol{equal}{*}

\begin{icmlauthorlist}
\icmlauthor{Rob Hesselink}{illc}
\icmlauthor{Wilker Aziz}{illc}
\end{icmlauthorlist}

\icmlaffiliation{illc}{Institute for Logic, Language, and Computation, University of Amsterdam, Netherlands}

\icmlcorrespondingauthor{Rob Hesselink}{rob.hesselink@pm.me}
\icmlcorrespondingauthor{Wilker Aziz}{w.aziz@uva.nl}

\icmlkeywords{deep latent variable models, normalising flows, discrete distributions}

\vskip 0.3in
]



\printAffiliationsAndNotice{}  

\begin{abstract}
Normalising flows (NFs) for discrete data are challenging because parameterising bijective transformations of discrete variables requires predicting discrete/integer parameters. Having a neural network architecture predict discrete parameters takes a non-differentiable activation function (\eg, the step function) which precludes gradient-based learning. To circumvent this non-differentiability, previous work has employed biased proxy gradients, such as the straight-through estimator. We present an unbiased alternative where rather than deterministically parameterising one transformation, we predict a distribution over latent transformations. With stochastic transformations, the marginal likelihood of the data is differentiable and gradient-based learning is possible via score function estimation. To test the viability of discrete-data NFs we investigate performance on binary MNIST. We observe great challenges with both deterministic proxy gradients and unbiased score function estimation. Whereas the former often fails to learn even a shallow transformation, the variance of the latter could not be sufficiently controlled to admit deeper NFs. 

\end{abstract}

\section{Introduction}
Normalising flows \citep{tabak2010density,rezende15NF} have been shown to be powerful density estimators for high-dimensional continuous data \cite{papamakarios2019normalizing}. For discrete data, NFs have received little attention thus far.  %
The foremost reason is the problem of gradient estimation of discrete functions. 
\citet{TranEtAl2019} and \citet{IntegerFlow} have resorted to a deterministic proxy gradient known as straight-through estimator \citep[STE;][]{bengio2013estimating}. 
In STE, ill-defined Jacobians are replaced by the identity matrix, effectively disregarding discontinuities in parameterisation. 
STE is a biased estimator and its effects as a function of the number of discrete variables (related to the dimensionality of the data) and/or the depth of the flow (a hyperparameter that controls expressiveness) are little understood.

To gain insight into the viability of discrete-data NFs, we propose to give transformations a latent treatment by means of turning their parameters into stochastic latent variables.
While this sacrifices tractable likelihood assessments, gradient-based learning is still possible via optimisation of the evidence lowerbound (ELBO), as in variational inference~\citep[VI;][]{Jordan+1999:VI}, and score function estimation~\citep[SFE;][]{rubinstein1976monte,paisley2012variational}. 
We use a generalised \textsc{xor} transformation \citep{tran2017hierarchical} to parameterise  autoregressive flows \citep{kingma2016improved,papamakarios2017masked} and compare the performance of STE and SFE to model binary data (MNIST).
Our experiments show that STE struggles in general and that SFE struggles with depth.\footnote{Code available on \href{https://github.com/RobDHess/Latent-DNFs}{github.com/robdhess/Latent-DNFs}} 

\section{Normalising Flows}

A normalising flow \citep[NF;][]{tabak2010density} is built upon an invertible and differentiable transformation of a continuous random variable with known density, i.e.
\begin{align}
    p_X(x) = p_Y(\inv{t}(x)) \djac{\inv{t}}{x}
\end{align}
where $\jac{\inv{t}}{x}$ is the matrix of partial derivatives of $\inv{t}$ assessed at $x$.
NFs have been used extensively in deep latent variable models \citep{rezende15NF,kingma2016improved}, where they play the role of a variational approximation to the model's true posterior, but also in density estimation \citep{dinh2014nice,dinh2017density,kingma2018glow,papamakarios2017masked}, where they enable exact likelihood-based learning with expressive distributions.

If $X$ takes on values in some \emph{discrete space}, a bijective mapping corresponds to an unambiguous relabelling of the sample space and thus incurs no distortion in volume (there is no notion of volume). The mass function of the transformed discrete variable is simply
\begin{align}
     p_X(x) = p_Y(\inv{t}(x)) ~ .
\end{align}
We can parameterise the transformation (either directly or its inverse) using a neural network---as long as we can guarantee the mapping to be bijective for any configuration of weights of the neural network.
\citet{TranEtAl2019} and \citet{IntegerFlow} have designed bijective transformations for discrete variables (the former focused on nominal variables, the latter on ordinal variables) used in a \emph{discrete-data normalising flow}. 
Richly parameterised transformations can correlate outcomes that are independent in the base leading to structured likelihoods for discrete data with applications such as text generation in natural language processing (e.g. language modelling, machine translation, text summarisation, dialogue modelling).\footnote{\citet{TranEtAl2019}, for example, design expressive non-autoregressive likelihoods for language problems, where autoregressiveness is the norm, and highlight efficiency as a motivation.}
The transformations underlying such discrete-data flows take \emph{discrete parameters} whose prediction by neural network architectures requires discontinuous activations with ill-defined derivatives (e.g. zero almost everywhere and sometimes undefined). 
\citet{TranEtAl2019} and \citet{IntegerFlow}  resort to a biased estimator known as straight-through estimator \citep[STE;][]{bengio2013estimating}. In particular, they use the deterministic version of STE, whereby one replaces the ill-defined Jacobian of a discontinuous activation by the identity matrix.

\section{Discrete-Data NFs}

Consider the case of binary data, e.g. $x \in \{0, 1\}^D$ and $y \in \{0, 1\}^D$, and, with no loss of generality, assume the base distribution is factorised:
\begin{align}
    p_{Y|\beta}(y) = \prod_{d=1}^D \Bern(y_d|\beta_d) ~.
\end{align}
An invertible transformation for binary data can be designed with the elementwise \textsc{xor} %
$\oplus$ \citep{TranEtAl2019}, namely,
\begin{align}
    x &=  \underbrace{y \oplus u}_{t(y; u)}  & \text{and} & & y &=  \underbrace{x \oplus u}_{\inv{t}(x; u)} ~,
\end{align}
where the flow parameters $u \in \{0, 1\}^D$ are themselves binary.
We can build an expressive flow, for example, by predicting $u$ with an autoregressive conditioner, \ie, $u=f(x; \theta)$ where $u_d$ depends on $x$ only through $x_{<d}$, and $\theta$ denotes the parameters of the conditioner.\footnote{For fixed $D$, this can be done with a feed-forward architecture such as MADE  \citep{germain2015made}, alternatively a recurrent or self-attention architecture can be used. Where autoregressiveness imposes a  bottleneck, a stack of bipartite transformations are an option \citep{dinh2017density}.}
Constraining $u$ to $\{0,1\}^D$ is exactly the source of non-differentiability that we address in this work.

\paragraph{Note on STE} Suppose $f(x; \theta) $ is a differentiable function from $\{0,1\}^D$ to $(0,1)^D$ (for example, a feed-forward NN with sigmoid outputs). An elementwise threshold function can constrain the outputs $ o = f(x; \theta)$ to $\{0,1\}^D$, i.e.
\begin{equation}
    h_d(o_d) = 
    \begin{cases}
    1 & \text{if } o_d > 0.5 \\
    0 & \text{otherwise}
    \end{cases}
\end{equation}
and we could design $u=h(f(x; \theta))$. However $\pdv{o_d}h(o)$ is undefined for $o_d=0.5$ and $0$ everywhere else. 
A model specified this way leads to a \emph{non-differentiable} likelihood, that is, for a given $x$, $p_{X|\theta,\beta}(x)$ is not a differentiable function of the parameters $\theta$. In the next section we will give $u$ stochastic treatment. Effectively, we specify a distribution $p_{XU|\theta, \beta}(x, u)$, differentiable with respect to its parameters, and above all, whose marginal $p_{X|\theta,\beta}(x)$ too is differentiable with respect to its parameters.
Note that STE ignores the non-differentiability of $h$ by defining 
$\jac{h}{o} = \mathbb I_D$
where $\mathbb I_D$ is the $D \times D$ identity matrix.

\section{Latent Transformations}

A straight-forward way to define a differentiable likelihood model that employs a discrete parameter $u$ is to sample the parameter from a distribution with discrete support and compute gradients \emph{in expectation}.
This is equivalent to thinking of the transformation as latent. 

We design a latent variable model
\begin{subequations}
\begin{align}
    p_{U|X,\theta}(u|x) &= \prod_{d=1}^{D} \Bern(u_d|f_d(x; \theta)) \label{eq:pu} \\
    p_{X|\beta,\theta}(x) &= \!\!\!\!\!\!\sum_{u \in \{0, 1\}^D} p_{Y|\beta}(\inv{t}(x; u))p_{U|X,\theta}(u|x) \label{eq:px}
\end{align}
\end{subequations}
where $f_d(x; \theta)$ depends on $x$ only through $x_{<d}$. 
Equation (\ref{eq:pu}) specifies a distribution over transformations, rather over the parameters of transformations. 
The marginalisation is clearly intractable---the sum ranges over $\mathcal O(2^D)$ outcomes---thus we approach this via variational inference \citep[VI;][]{Jordan+1999:VI} with an approximate posterior
\begin{align}
    q_{U|X,\lambda}(u|x) &= \prod_{d=1}^D \Bern(u_d|g_d(x; \lambda)) ~, \label{eq:qu}
\end{align}
where, unlike $f(x; \theta)$, $g(x; \lambda)$ need not be  auto-regressive. 
We then optimise parameters using gradient estimates of the evidence lowerbound (ELBO) with respect to $\theta$ and $\lambda$ (and possibly $\beta$ too, though we typically leave the base fixed):
\begin{align}\label{ELBO}
    &\log p_{X|\beta,\theta}(x) \ge \mathbb E\left[ \log \frac{ p_{Y|\beta}(\inv{t}(x; u))p_{U|X,\theta}(u|x) }{q_{U|X,\lambda}(u|x)} \right] ~,
\end{align}
where the expectation is taken \wrt $q_{U|X,\lambda}(u|x)$.
This model learns a distribution over latent transformations, from which a sample parameterises a discrete-data normalising flow. 
Clearly, unlike a standard normalising flow, we do not have exact marginal likelihood assessments, but we can estimate a lowerbound via importance sampling (this is the common practice with VAEs). NFs have been used in VAEs, but usually to model the latent space \citep{rezende15NF}, in this proposal we use VI to train a model whose sampling distribution is itself an NF. 
We do maintain tractability of sampling though.

\paragraph{Multiple layers}
In general, we have a sequence of $L$ transformations $u = \langle u^{(1)}, \ldots, u^{(L)}\rangle$, each $u^{(\ell)} \in \{0,1\}^D$. We then obtain a base sample $y=t(x; u)$ by iteratively transforming the data sample: $x^{(\ell)} = x^{(\ell-1)} \oplus u^{(\ell)}$ for $\ell=1,\ldots,L$, where $x=x^{(0)}$ and $y=x^{(L)}$ and with $\oplus$ applying elementwise. 
To model with multiple flows we make a conditional independence assumption, namely, that $u^{(\ell)}$ depends only on $u^{(\ell-1)}$, both in the generative model $p_{U|X,\theta}(u|x) = \prod_{\ell=1}^L  p(u^{(\ell)}|u^{(\ell-1)}, \theta_\ell)$, and in the approximate posterior $q_{U|X,\lambda}(u|x) = \prod_{\ell=1}^L q(u^{(\ell)}|u^{(\ell-1)}, \lambda_\ell)$, where $u^{(0)} \coloneqq x$ and we omit subscripts to avoid clutter. Appendix \ref{sec:Model++} contains more information about the factorisation.

\paragraph{Other latent transformation NFs} 
Marginalisation of latent transformations has been proposed as a means to increase flexibility of continuous-data NFs, which struggle to accommodate topological differences between the target and the base distribution.  \citet{dinh2019rad} introduce a finite mixture of transformations, where marginalisation is possible if the number of components is small. For increased expressiveness, \citet{cornish2019localised} propose  continuously indexed NFs (\ie, a compound distribution whose sampling distribution is an NF) and approach approximate marginalisation via variational inference. 
Rather than (exactly or approximately) marginalising out members of a parametric family of transformations, one can realise the mapping between target and base distributions as a stochastic process~\citep{pmlr-v37-sohl-dickstein15,wu2020stochastic}. %

\section{Gradient Estimation}

The intractability of the marginal likelihood extends to the ELBO and for that reason we must resort to Monte Carlo (MC) gradient estimation. 
The gradient of the ELBO \wrt to the parameters $\beta$ of the base distribution is straightforward to estimate using a sample $u \sim q_{U|X=x,\lambda}$,  $\grad_{\beta} \ELBO \overset{\text{MC}}{\approx} \grad_{\beta} \log p_{Y|\beta}(t(x; u))$. That is because $\beta$ is not involved in the sampling of $u$.
Similarly, estimating the gradient \wrt the parameters $\theta$ of the generative model poses no challenge,  $\grad_{\theta} \ELBO \overset{\text{MC}}{\approx} \grad_{\theta} \log p_{U|X,\theta}(u|x)$.
The gradient \wrt the parameters $\lambda$ of the approximate posterior, $\grad_{\lambda} \ELBO =$
\begin{equation}
    \grad_\lambda \mathbb E\left[ \log p_{Y|\beta}(t(x; u))\right] -\grad_\lambda \KL(q_{U|x,\lambda} || p_{U|x,\theta}) \label{eq:gradlamb}
\end{equation}
is less trivial to estimate since $\lambda$ is involved in sampling $u$, but it can be rewritten as follows,
\begin{subequations}
\begin{align}
    &\mathbb E\left[ \log p_{Y|\beta}(t(x; u))\grad_{\lambda} \log q_{U|X,\lambda}(u|x)\right] \label{eq:SFEbase} \\
    &-\mathbb E\left[ \log \frac{p_{U|X,\theta}(u|x)}{q_{U|X,\lambda}(u|x)} \grad_\lambda \log q_{U|X,\lambda}(u|x) \right] ~, \label{eq:SFEKL}
\end{align}
\end{subequations}
for which MC estimation is possible. The result is the SFE, which is generally very noisy.
For variance reduction we follow the steps of \citet{mnih2014neural} closely.
That is, to reduce variance of the first term (\ref{eq:SFEbase}), we subtract from the learning signal a self-critic \citep{rennie2017self} and standardise the result using moving estimates of mean and standard deviation.\footnote{A self-critic here corresponds to $\log p_{Y|\beta}(t(x; u'))$ for an independently sampled transformation $u'$.}
These baselines can be defined for each observed variable since $\log p_{Y|\beta}(t(x; u)) = \sum_{d=1}^D \log p_{Y|\beta}(x_d \oplus u_d^{(1)} \ldots \oplus u_d^{(L)})$.
For the second term (\ref{eq:SFEKL}), in addition to using baselines, we rewrite the SFE factoring out of the computation as many terms as possible in order to reduce variance. 

\paragraph{Local KL} The gradient of the KL term is expressed in terms of nested expectations, and due to conditional independence assumptions in the model and in the posterior approximation, we can improve upon a vanilla SFE in manner reminiscent of how \citet{mnih2014neural} improved the SFE for hidden layers with binary hidden units.
The first-order Markov assumption we make over latent parameters implies weighing the score of $u^{(\ell)}$ by the sum of log-ratios due to two KL terms: the one \wrt the distribution of $U^{(\ell)}$ and the one \wrt the distribution of $U^{(\ell+1)}$. %
Thus, we can rewrite the SFE for the KL term (complete derivation in Appendix~\ref{sec:SFE-KL}). 
For each layer $\ell$, $\grad_{\lambda_\ell} \KL \approx$
\begin{subequations}
\begin{align}
    &\log \frac{p(u^{(\ell)}|u^{(\ell-1)}, \theta_\ell)}{q(u^{(\ell)}| u^{(\ell-1)}, \lambda_\ell)} \grad_{\lambda_\ell} \log q(u^{(\ell)}|u^{(\ell-1)}, \lambda_\ell) \label{eq:KLell}\\
    +&\log \frac{p(u^{(\ell+1)}|u^{(\ell)}, \theta_\ell)}{q(u^{(\ell+1)}| u^{(\ell)}, \lambda_\ell)} \grad_{\lambda_\ell} \log q(u^{(\ell)}|u^{(\ell-1)}, \lambda_\ell) \label{eq:KLell+1} ~,
\end{align}
\end{subequations}
where $\lambda_\ell$ corresponds to the parameters of the neural network $g$ that maps from $u^{(\ell-1)}$ to a distribution over possible assignments to the $\ell$th transformation.\footnote{We could potentially estimate the two terms independently, which then allows scaling the second SFE by $\gamma > 0$.}
Finally, (\ref{eq:KLell}) can be further improved 
\begin{subequations}\label{eq:betterKLell}
\begin{align}
    &\grad_{\lambda_\ell}\sum_{d=1}^D \mathbb H(U_d^{(\ell)}|u^{(\ell-1)}, \lambda_\ell) \label{eq:betterKLell-H} \\
    +&\sum_{d=1}^D \log p(u_{\ge d}^{(\ell)}| u^{(\ell-1)}, u_{<d}^{(\ell)})  \grad_{\lambda_\ell} \log q(u^{(\ell)}|u^{(\ell-1)}, \lambda_\ell)  \label{eq:betterKLell-SFE} 
\end{align}
\end{subequations}
by factoring out the closed-form entropy of and realising that $u_d^{(\ell)}$ cannot affect $\log p(u_{< d}^{(\ell)}|u^{(\ell-1)}, \theta_\ell)$.
For further variance reduction of the SFE in (\ref{eq:betterKLell-SFE}),  we can again employ a self-critic and standardisation of the learning signal, though this time, these techniques are applied independently for each layer.

\paragraph{Special case}
Since including both a variational posterior and an autoregressive generative model can be computationally costly, we also consider the situation where we choose $q_{U|X,\lambda}(u|x) \coloneqq p_{U|X,\theta}(u|x)$. This corresponds to a simpler form of VI where rather than introducing an independent posterior approximation we estimate a lowerbound on log-likelihood by sampling from the generative model directly. This leads to KL  evaluating to zero, and makes gradient estimation \wrt $\theta$ a matter of score function estimation. In this case, we use the bound $\log p_{X|\beta,\theta}(x) \ge$
\begin{equation} \label{eq:reinforcement}
    \mathcal L = \mathbb E_{p_{U|X,\theta}(u|x)}\left[ \log p_{Y|\beta}(\inv{t}(x; u)) \right] ~,
\end{equation}
While the gradient \wrt $\beta$ is unchanged, the gradient \wrt $\theta$ can be split in two terms. For this, we group the $L$ parameters $\langle u^{(1)}_d, \ldots, u^{(L)}_d \rangle$ that transform the $d$th observation, and denote them simply by $u_d$. Then $\grad_\theta \mathcal L =$
\begin{subequations}
\begin{align}
    &\sum_{d=1}^D \mathbb E_{} \left[r(u_d) \grad_{\theta}\log p(u_d|u_{<d}) + r(u_d) \grad_{\theta} \log p(u_{<d})  \right]
\end{align}
\end{subequations}
with expectations taken \wrt $p_{U|X,\theta}(u|x)$, and  $r(u_d) \coloneqq \log p_{Y|\beta}(x_d \oplus u^{(1)}_d \ldots \oplus u^{(L)}_d)$.
For the full derivation, see appendix \ref{sec:special case}. Note how the reward for the $d$th observation scales the score of its transformation (first term) as well as the score of the transformations of preceding observations (second term). We want to note similarities to the Bellman equation, where rewards are discounted for past actions. If we estimate the two terms independently, that is, using two independent samples $u$ and $u'$, stochastic gradient optimisation allows us to choose different scaling constants for each term. 
We opt for an impatient approach and set the constant for the second term negligibly small.

\section{Experiments}
We compare SFE and STE on binarised MNIST. We binarise stochastically by interpreting the pixel intensities as Bernoulli probabilities. Every layer uses a 784-dimensional MADE \cite{germain2015made} and models are trained using Adam with learning rate 1e-3. The base distribution is fixed with Bernoulli parameters $[0.9, 0.1]$. Stochastic models are evaluated using $1,000$ samples, and we also report the performance of a single flow, namely the one specified by a greedy approximation to $\argmax_u q_{U|X,\lambda}(u|x)$. 

\begin{table}[h]
    \centering
    \begin{tabular}{l r r r}
        \toprule
        {\bf Model}          & Depth                & {\bf NLL} $\downarrow$ & Greedy  \\ \midrule
        STE           & 1            & $222.1$          \\
                      & 2         &  $211.2$           \\
                      & 4            &  $205.6$           \\
                      & 8         &  $209.1$          \\ \midrule
        SFE (full posterior) &  &  &  \\
        ~ greedy self-critic & 1                &    $293.7$     & $205.3$ \\
        ~ running average & 1       &    $242.4$ & $207.3$       \\
        \bottomrule
        SFE (special case) &  &   &  \\
        ~ greedy self-critic & 1                &    $267.9$     & $267.7$ \\
        ~ running average & 1       &    $190.1$ & $189.6$       \\
        \bottomrule
\end{tabular}
    \caption{Negative log-likelihood on binarised MNIST. Results show best of three independent runs.}
    \label{tab:likelihood}
\end{table}

Table \ref{tab:likelihood} shows that by applying proper variance reduction techniques, SFE can outperform STE, even for architectures with fewer parameters. In the special case where $q_{U|X,\lambda}(u|x) \coloneqq p_{U|X,\theta}(u|x)$ performance is highest, possibly due to lower gradient variance.  We did find, however, that performance did not increase with depth. Multiple stochastic layers increases variance in performance and the model encourages deeper layers to become fixed. Samples of SFE and STE can be found in appendix \ref{sec:samples}.

We want to emphasize the possibility of treating the flow stochastically during training to compute the gradients, but reverting to a deterministic forward pass after that. By doing this, we train using unbiased estimates of the gradient, without losing exact likelihood assessment.



\section{Conclusion}
We have presented a new technique for training discrete normalising flows by treating the transformation parameters as latent variables. Our unbiased gradients lead to better performance on binarised MNIST compared to straight-through, while we sacrifice exact likelihood assessment. Further work could experiment with more powerful architectures and recent developments in discrete gradient estimation, such as \cite{RELAX}. 

\section*{Acknowledgements} 

This project is supported by the European Union's Horizon 2020 research and innovation programme under grant agreement No 825299 (Gourmet). We thank Emiel Hoogeboom for insightful discussions.  

\bibliography{latentflow/BIB}
\bibliographystyle{icml2020}

\onecolumn
\appendix

\section{\label{sec:Model++}Model}

\subsection{Specification}

\paragraph{Generative model} We make a first-order Markov assumption across layers and model the units within a layer autoregressively:
\begin{align}
    p_{U|X,\theta}(u|x) &= \prod_{\ell=1}^L  p(u^{(\ell)}|u^{(\ell-1)}, \theta_\ell)\\
    p(u^{(\ell)}|u^{(\ell-1)}, \theta_\ell) &= \prod_{d=1}^D p(u_d^{(\ell)}|u^{(\ell-1)}, u_{<d}^{(\ell)}, \theta_\ell) \\
    U_d^{(\ell)}|\theta_\ell, u^{(\ell-1)}, u_{<d}^{(\ell)} &\sim \Bern(f_d(u^{(\ell-1)}, u^{(\ell)}; \theta_\ell)) ~,
\end{align}
where $u^{(0)} \coloneqq x$ and $f_d(\cdot; \theta_\ell)$ depends on $u^{(\ell)}$ only through $u_{<d}^{(\ell)}$.

\paragraph{Approximate posterior} We make a first-order Markov assumption across layers and mean field assumption within layers:
\begin{align}
    q_{U|X,\lambda}(u|x) &= \prod_{\ell=1}^L q(u^{(\ell)}|u^{(\ell-1)}, \lambda_\ell) \\
    q(u^{(\ell)}|u^{(\ell-1)}, \lambda_\ell) &= \prod_{d=1}^D q(u_d^{(\ell)}|u^{(\ell-1)}, \lambda_\ell) \\ 
    U_d^{(\ell)}|\lambda_\ell, u^{(\ell-1)} &\sim  \Bern(g_d(u^{(\ell-1)}; \lambda_\ell)) ~,
\end{align}
where $u^{(0)} = x$. Here $g_d(\cdot; \lambda_\ell)$ conditions freely on all of its inputs.

\paragraph{ELBO}
\begin{align}
    \mathbb E_{q_{U|X,\lambda}(u|x)}\left[ \log p_{Y|\beta}(\inv{t}(x; u)) + \log\frac{p_{U|X,\theta}(u|x)}{q_{U|X,\lambda}(u|x)}\right] = \mathbb E_{q_{U|X,\lambda}(u|x)}\left[ \log p_{Y|\beta}(\inv{t}(x; u))\right] - \KL(q_{U|x,\lambda} || p_{U|x,\theta})
\end{align}

\subsection{Gradient estimation}
Unless otherwise noted, we derive MC estimates on a single sample $u \sim q_{U|X=x,\lambda}$.
\paragraph{Base distribution} 
\begin{equation}
    \grad_{\beta} \ELBO \overset{\text{MC}}{\approx} \grad_{\beta} \log p_{Y|\beta}(\inv{t}(x; u))~.
\end{equation}

\paragraph{Generative model} 
\begin{equation}
    \grad_{\theta} \ELBO \overset{\text{MC}}{\approx} \grad_{\theta} \log p_{U|X,\theta}(u|x)~.
\end{equation}

\paragraph{Approximate posterior} 
\begin{align}
    \grad_{\lambda}  \ELBO &\overset{\text{MC}}{\approx} \grad_\lambda \mathbb E_{q_{U|X,\lambda}(u|x)}\left[ \log p_{Y|\beta}(\inv{t}(x; u))\right] -\grad_\lambda \KL(q_{U|X=x,\lambda} || p_{U|X=x,\theta})
\end{align}
For the first term we employ the score function estimator (SFE):
\begin{equation}
    \mathbb E_{q_{U|X,\lambda}(u|x)}\left[ \log p_{Y|\beta}(\inv{t}(x; u))\grad_{\lambda} \log q_{U|X,\lambda}(u|x)\right] ~.
\end{equation}
For variance reduction we subtract a self-critic, namely, $\log p_{Y|\beta}(\inv{t}(x; u'))$ for an independent sample $u'$, and standardise the result using moving estimates of mean and standard deviation. These baselines are applied per pixel since the reward factorises: $\log p_{Y|\beta}(\inv{t}(x; u)) = \sum_{d=1}^D \log p_{Y|\beta}(x_d \oplus u^{(1)}_d \oplus \ldots \oplus y^{(L)}_d)$.
The second term is made of nested KL terms, and due to conditional independence assumptions in the model and in the approximation, we can improve upon a vanilla SFE (see next).

\section{\label{sec:SFE-KL}Gradient of KL}

We make a conditional independence assumption, namely, $u^{(\ell)}$ is independent of all but the previous assignment $u^{(\ell-1)}$, which can be exploited to rewrite the SFE for the KL term.
We are looking to simplify
\begin{align}
    \grad_\lambda \KL(q_{U|x,\lambda} || p_{U|x,\theta}) = \mathbb E_{q_{U|X,\lambda}(u|x)}\left[ \log\frac{p_{U|X,\theta}(u|x)}{q_{U|X,\lambda}(u|x)} \grad_\lambda \log q_{U|X,\lambda}(u|x) \right]~.
\end{align}
Define a ``local reward'' as
\begin{align}
    r(u^{(\ell)}, u^{(\ell-1)}) \coloneqq \sum_{d=1}^D \log \frac{p(u^{(\ell)}_d| u^{(\ell-1)}, u^{(\ell)}_{<d})}{q(u^{(\ell)}_d|u^{(\ell-1)}, \lambda_\ell)} ~.
\end{align}
Let expectations be expressed \wrt $q_{U|X,\lambda}$, then the gradient of the KL with respect to the parameters of the $\ell$th layer is:
\begin{subequations}
\begin{align}
    \grad_{\lambda_\ell} \KL =&\mathbb E\left[ \left(\sum_{k=1}^L r(u^{(k)}, u^{(k-1)})\right) \grad_{\lambda_\ell} \log q_{U|X,\lambda}(u|x) \right] \\
    \intertext{Note that we only need the $\ell$th score.}
    =&\mathbb E\left[ \left(\sum_{k=1}^L r(u^{(k)}, u^{(k-1)})\right) \grad_{\lambda_\ell} \log q(u^{(\ell)}|u^{(\ell-1)}, \lambda_\ell) \right] \\
    \intertext{And that we can expand the reward explicitly to identify constant terms.}
    =&\mathbb E\left[ \underbrace{\left(\sum_{k=1}^{\ell-1} r(u^{(k)}, u^{(k-1)})\right)}_{\text{constant w.r.t. }u^{(\ell)}} \grad_{\lambda_\ell} \log q(u^{(\ell)}|u^{(\ell-1)}, \lambda_\ell) \right] \\
    +&\mathbb E\left[ \underbrace{\left(\sum_{k=\ell+2}^{L} r(u^{(k)}, u^{(k-1)})\right)}_{\text{constant w.r.t. }u^{(\ell)}} \grad_{\lambda_\ell} \log q(u^{(\ell)}|u^{(\ell-1)}, \lambda_\ell) \right] \\
    +&\mathbb E\left[ \underbrace{\left(r(u^{(\ell)}, u^{(\ell-1)}) + r(u^{(\ell+1)}, u^{(\ell)})\right)}_{\text{local rewards}} \grad_{\lambda_\ell} \log q(u^{(\ell)}|u^{(\ell-1)}, \lambda_\ell) \right] ~,
    \intertext{Recall that the expected value of the score is $0$, \ie, $\mathbb E_{q(u^{(\ell)}|u^{(\ell-1)},\lambda_\ell)}\left[  \grad_{\lambda_\ell} \log q(u^{(\ell)}|u^{(\ell-1)},\lambda_\ell) \right] = 0$, thus the expectations involving rewards which are constant w.r.t. $u^{(\ell)}$ disappear. This allows us to simplify the expression keeping only the part that includes the local rewards:}
    =&\mathbb E\left[ \left(r(u^{(\ell)}, u^{(\ell-1)}) + r(u^{(\ell+1)}, u^{(\ell)})\right) \grad_{\lambda_\ell} \log q(u^{(\ell)}|u^{(\ell-1)},\lambda_\ell) \right] ~.
\end{align}
\end{subequations}
This estimator can be further refined. In the part that accounts for $r(u^{(\ell)}, u^{(\ell-1)})$ we can solve the entropy term exactly:
\begin{subequations}
\begin{align}
    &\mathbb E_{q(u^{(\ell)}|u^{(\ell-1)},\lambda_\ell)}\left[ r(u^{(\ell)}, u^{(\ell-1)}) \grad_{\lambda_\ell} \log q(u^{(\ell)}|u^{(\ell-1)},\lambda_\ell) \right] \\
    &=\mathbb E_{q(u^{(\ell)}|u^{(\ell-1)},\lambda_\ell)}\left[ \log \frac{p(u^{(\ell)}|u^{(\ell-1)}, \theta_\ell)}{q(u^{(\ell)}|u^{(\ell-1)}, \lambda_\ell)} \grad_{\lambda_\ell} \log q(u^{(\ell)}|u^{(\ell-1)}, \lambda_\ell) \right] \\
    &=\mathbb E_{q(u^{(\ell)}|u^{(\ell-1)}, \lambda_\ell)}\left[ \log p(u^{(\ell)}| u^{(\ell-1)}, \theta_\ell)  \grad_{\lambda_\ell} \log q(u^{(\ell)}|u^{(\ell-1)}, \lambda_\ell) \right] \\
    &- \underbrace{\mathbb E_{q(u^{(\ell)}|u^{(\ell-1)},\lambda_\ell)}\left[\log q(u^{(\ell)}|u^{(\ell-1)},\lambda_\ell) \grad_{\lambda_\ell} \log q(u^{(\ell)}|u^{(\ell-1)},\lambda_\ell) \right]}_{\grad_{\lambda_\ell} \mathbb H(U^{(\ell)}|u^{(\ell-1)}, \lambda_\ell))} ~.
\end{align}
\end{subequations}
Recall that with a mean field assumption the entropy can be computed exactly
\begin{align}
    \mathbb H(U^{(\ell)}|u^{(\ell-1)}, \lambda_\ell)) &= \sum_{d=1}^D \mathbb H(q(u_d^{(\ell)}|u^{(\ell-1)}, \lambda_\ell)) ~.
\end{align}
The negative cross entropy term can not be computed exactly because the generative model is autoregressive, that is, $u_d^{(\ell)}$ depends on $u^{(\ell)}_{<d}$. 
Still, its SFE can be further simplified. The argument is once again based on iterated expectations:
\begin{subequations}
\begin{align}
    &\mathbb E\left[ \log p(u^{(\ell)}|u^{(\ell-1)},\theta_\ell)  \grad_{\lambda_\ell} \log q(u^{(\ell)}|u^{(\ell-1)},\lambda_\ell) \right] \\
    &= \sum_{d=1}^D \mathbb E\left[ \underbrace{\left(\sum_{k=1}^{d-1} \log p(u_k^{(\ell)}|u^{(\ell-1)}, u_{<k}^{(\ell)}, \theta_\ell) \right)}_{\text{constant w.r.t }u_d^{(\ell)}}  \grad_{\lambda_\ell} \log q(u^{(\ell)}|u^{(\ell-1)},\lambda_\ell) \right] \\
    &+ \sum_{d=1}^D \mathbb E\left[ \left(\sum_{k=d}^{D} \log p(u_k^{(\ell)}|u^{(\ell-1)}, u_{<k}^{(\ell)}, \theta_\ell) \right)  \grad_{\lambda_\ell} \log q(u^{(\ell)}|u^{(\ell-1)},\lambda_\ell) \right] \\
    &= \sum_{d=1}^D \mathbb E\left[ \left(\sum_{k=d}^{D} \log p(u_k^{(\ell)}|u^{(\ell-1)}, u_{<k}^{(\ell)}, \theta_\ell) \right)  \grad_{\lambda_\ell} \log q(u^{(\ell)}|u^{(\ell-1)},\lambda_\ell) \right]
\end{align}
\end{subequations}

\section{\label{sec:special case}Special case}
Here we lowerbound the marginal log-likelihood of the data by direct application of Jensen's inequality without an independent variational approximation. 
We are interested in the gradient 
\begin{equation}
    \grad_{\theta} \mathbb E_{p_{U|X,\theta}(u|x)}\left[ \log p_{Y|\beta}(\inv{t}(x; u)) \right]
\end{equation}

We can first express the gradient for layer $\ell$ as 
\begin{subequations}
\begin{align}
&\grad_{\theta_\ell}\sum_{\ell=1}^L  \mathbb E_{p(u^{(<\ell)}|\theta)} \mathbb E_{p(u^{(\ell)}|u^{(\ell-1)}, \theta_\ell)} \left[\log p_{Y|\beta}(\inv{t}(x; u))\right] \\
=&\grad_{\theta_\ell}\sum_{\ell=1}^L  \mathbb E_{p(u^{(<\ell)}|\theta)}  
\sum_{d=1}^D \mathbb E_{p(u^{(\ell)}_{<d}|u^{(\ell-1)}, \theta_\ell)}
\mathbb E_{p(u^{(\ell)}_d|u^{(\ell)}_{<d}, u^{(\ell-1)}, \theta_\ell)}
\left[\log p_{Y|\beta}(\inv{t}(x; u))  \right]
\end{align}
\end{subequations}
Where we first applied the Markov property and secondly iterated expectation within the layer. If we introduce a proposal distribution $\rho(u^{(\ell)}_{<d})$, independent of $\theta_\ell$, and define 
\begin{align}
    \omega(u^{(\ell)}_{<d}, \theta) &= \frac{p_\theta(u^{(\ell)}_{<d})}{\rho(u^{(\ell)}_{<d})} \\
    \mathcal L(u^{(\ell)}_{<d}, \theta) &= \mathbb E_{p(u^{(\ell)}_d|u^{(\ell)}_{<d}, u^{(\ell-1)}, \theta_\ell)}
\left[\log p_{Y|\beta}(\inv{t}(x; u))  \right]
\end{align}
then $\grad_{\theta_\ell}  \sum_{d=1}^D \mathbb E_{p(u^{(\ell)}_{<d}|u^{(\ell-1)}, \theta_\ell)} \left[ \mathcal L(u^{(\ell)}_{<d}, \theta_\ell) \right]$
\begin{subequations}
\begin{align}
    &= \grad_{\theta_{\ell}}  \sum_{d=1}^D \mathbb E_{\rho(u^{(\ell)}_{<d})}\left[ \omega(u^{(\ell)}_{<d}, \theta_\ell)\mathcal L(u^{(\ell)}_{<d}, \theta_\ell) \right]  \\
    &= \sum_{d=1}^D \mathbb E_{\rho(u_{<d})}\left[\omega(u^{(\ell)}_{<d}, \theta_\ell)  \grad_{\theta_\ell} \mathcal L(u^{(\ell)}_{<d}, \theta_\ell) +
      \mathcal L(u^{(\ell)}_{<d}, \theta_\ell)  \grad_{\theta_\ell} \omega(u^{(\ell)}_{<d}, \theta_\ell) \right]  ~.
\end{align}
\end{subequations}
Note that the first term inside the expectation can be estimated via the score function estimator i.e. 
\begin{align}
    \omega(u^{(\ell)}_{<d}, \theta_\ell)  \grad_{\theta_\ell} \mathcal L(u^{(\ell)}_{<d}, \theta_\ell) &= \omega(u^{(\ell)}_{<d}, \theta_\ell) \mathbb E_{p(u^{(\ell)}_d|u^{(\ell)}_{<d}, u^{(\ell-1)}, \theta_\ell)} \left[ \log p_{Y|\beta}(\inv{t}(x; u)) \grad_{\theta_\ell} p(u^{(\ell)}_d|u^{(\ell)}_{<d}, u^{(\ell-1)}, \theta_\ell) \right] ~ .
\end{align}
For the second term we can write 
\begin{subequations}
\begin{align}
    \mathcal L(u^{(\ell)}_{<d}, \theta_\ell)  \grad_{\theta_\ell} \omega(u^{(\ell)}_{<d}, \theta_\ell) &= \frac{\mathcal L(u^{(\ell)}_{<d}, \theta_\ell) }{\rho(u^{(\ell)}_{<d})} \grad_{\theta_\ell} p(u^{(\ell)}_{<d}|u^{(\ell-1)}, \theta_\ell) \\ 
    &= \omega(u^{(\ell)}_{<d}, \theta_\ell) \mathcal L(u^{(\ell)}_{<d}, \theta_\ell) \grad_{\theta_\ell} p(u^{(\ell)}_{<d}|u^{(\ell-1)}, \theta_\ell)  ~.
\end{align}
\end{subequations}
Where our proposal is $\rho(u^{(\ell)}_{<d}) \coloneqq p(u^{(\ell)}_{<d}|u^{(\ell-1)}, \theta_\ell)$, we have $\omega(u^{(\ell)}_{<d}, \theta_\ell)  = 1$. 
This means that the first term becomes a local learning signal
\begin{align}
     \sum_{\ell=1}^L  &\mathbb E_{p(u^{(<\ell)}|\theta)}\mathbb E_{p(u^{(\ell)}_d|u^{(\ell)}_{<d}, u^{(\ell-1)}, \theta_\ell)} \left[ \log p_{Y|\beta}(\inv{t}(x; u)) \grad_{\theta_\ell} p(u^{(\ell)}_d|u^{(\ell)}_{<d}, u^{(\ell-1)}, \theta_\ell) \right],
\end{align}
where the learning signal derived from the $d$th observation interacts directly with the gradient of the log-probability of the $d$th latent variable.
The second term becomes 
\begin{align}
    \sum_{\ell=1}^L  &\mathbb E_{p(u^{(<\ell)}|\theta)}\mathbb E_{p(u^{(\ell)}_d|u^{(\ell)}_{<d}, u^{(\ell-1)}, \theta_\ell)} \left[ \log p_{Y|\beta}(\inv{t}(x; u)) \grad_{\theta_\ell} p(u^{(\ell)}_{<d}|u^{(\ell-1)}, \theta_\ell)\right]
\end{align}
where the learning signal derived from the $d$th observation interacts with the gradient of the log-probability of the $d$th latent prefix. 

\section{\label{sec:samples}Samples}

\begin{figure}[h]
    \centering
    \includegraphics[width=0.7\textwidth]{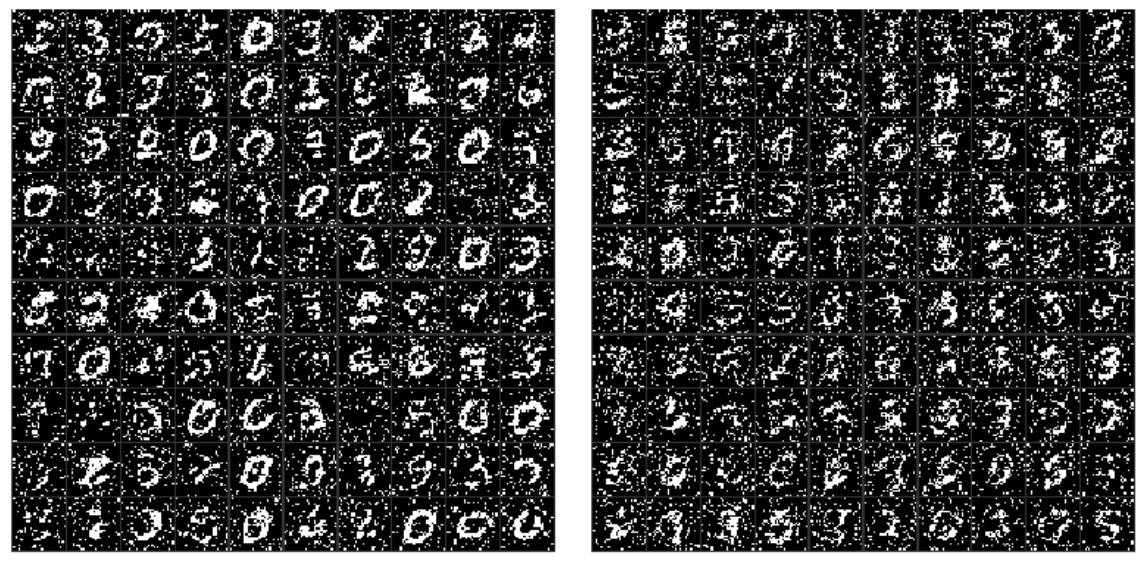}
    \caption{100 Samples from the best performing SFE (left) and best performing STE (right)}
    \label{fig:samples}
\end{figure}

\end{document}